\pdfoutput=1

\documentclass[11pt]{article}

\usepackage[final]{acl}

\usepackage{times}
\usepackage{latexsym}
\usepackage{booktabs}

\usepackage[T1]{fontenc}

\usepackage[utf8]{inputenc}
\usepackage{subcaption}

\usepackage{microtype}

\usepackage{inconsolata}

\usepackage{graphicx}

\usepackage{xcolor}

\title{Referring Expressions as a Lens into Spatial Language Grounding in Vision-Language Models}

\author{Akshar Tumu \\
  UC San Diego / \\ La Jolla, CA, USA \\
  \texttt{atumu@ucsd.edu} \\\And
  Varad Shinde \\
  IIT Kanpur / \\ Kanpur, Uttar Pradesh, India\\
  \texttt{varadshinde@iitbhilai.ac.in} \\\And
  Parisa Kordjamshidi \\
  Michigan State University / \\ East Lansing, MI, USA\\
  \texttt{kordjams@msu.edu} \\
  }

\begin{document}
\maketitle
\begin{abstract}
Spatial Reasoning is an important component of human cognition and is an area in which the latest Vision-language models (VLMs) show signs of difficulty. The current analysis works use image captioning tasks and visual question answering. In this work, we propose using the Referring Expression Comprehension task instead as a platform for the evaluation of spatial reasoning by VLMs. This platform provides the opportunity for a deeper analysis of spatial comprehension and grounding abilities when there is 1) ambiguity in object detection, 2) complex spatial expressions with a longer sentence structure and multiple spatial relations, and 3) expressions with negation (‘not’). In our analysis, we use task-specific architectures as well as large VLMs and highlight their strengths and weaknesses in dealing with these specific situations. While all these models face challenges with the task at hand, the relative behaviors depend on the underlying models and the specific categories of spatial semantics (topological, directional, proximal, etc.). Our results highlight these challenges and behaviors and provide insight into research gaps and future directions.
\end{abstract}

\section{Introduction}\label{sec:intro}
Vision-language model (VLM) research has boomed in the recent past, owing to the enhanced user interaction and accessibility they provide. Models such as GPT 4o\footnote{\url{https://openai.com/index/hello-gpt-4o/}}, LLaVA (\citealp{llava}), Google Gemini (\citealp{gemini}) have become adept at solving vision-language tasks such as Visual Question Answering (VQA), Image Captioning, and more. However, VLMs still lack human-level ‘Spatial Reasoning' capabilities (\citealp{vsr,whatsup}). Spatial reasoning involves comprehending relations that depict the absolute/relative position or orientation of an object, such as ‘left’, ‘above’, or ‘near’. Inaccurate spatial reasoning by VLMs can lead to serious consequences in embodied AI domains such as autonomous driving and surgical robotics. A focused analysis of VLMs' spatial reasoning capabilities can help identify potential reasoning issues.

Most of the previous works confine their analysis to testing which models work well for spatial relations. We go further to analyze the comparative performance of these models for spatial categories that represent different orientational and positional relations between objects. A novel aspect of our work is the analysis of the effect of varying spatial composition (number of spatial relations) in the expressions on the performance of the models.

Previous works focused on spatial analysis with image captioning-related tasks, thus failing to locate the source of error in the presence of visual and linguistic ambiguity. To avoid this, we adopt the Referring Expression Comprehension (REC) task (\citealp{qiao2020referring}) where the models output bounding boxes around the target entity based on a natural language expression, the analysis of which could reveal the parts of the input that the models fail to comprehend. Comprehension accuracy (or simply, accuracy) is a common metric for this task; it captures how often a model correctly outputs the bounding box around the target entity.


We test four popular VLMs - LLaVA (\citealp{llava}), Grounding DINO (GDINO) (\citealp{gdino}), DeepSeek-VL2 (\citealp{deepseek}), and Qwen2.5-VL (\citealp{qwen}). We also include ‘MGA-Net’ (\citealp{mganet}), a model specifically designed for the REC task. These models offer diversity in the evaluation as they cover different architectural elements, training strategies, and input formats. We further compare these models with an object detector baseline to test if the images are truly complex and require elaborate referring expressions to ground the correct object.
\begin{figure*}[!tb]
\centering
\begin{subfigure}[t]{0.16\linewidth}
  \includegraphics[width=\linewidth,height=2.5cm]{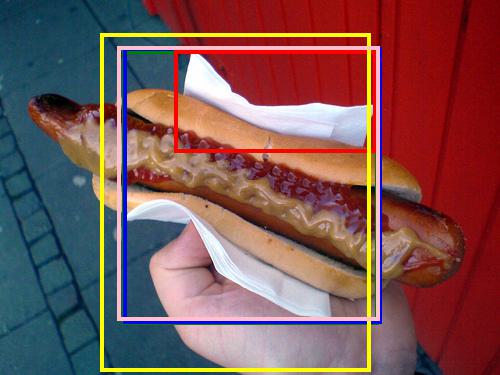}
  \caption{The white napkin that is wrapped around the hot dog}
  \label{fig:1a}
\end{subfigure} \hfill
\begin{subfigure}[t]{0.17\linewidth}
  \includegraphics[width=\linewidth,height=2.5cm]{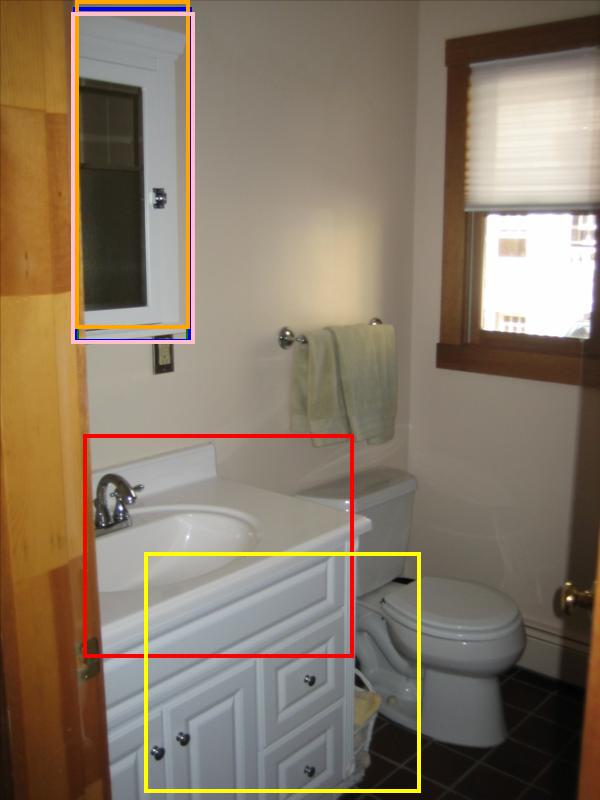}
  \caption{The white box that is around the mirror}
  \label{fig:1b}
\end{subfigure} \hfill
\begin{subfigure}[t]{0.17\linewidth}
  \includegraphics[width=\linewidth,height=2.5cm]{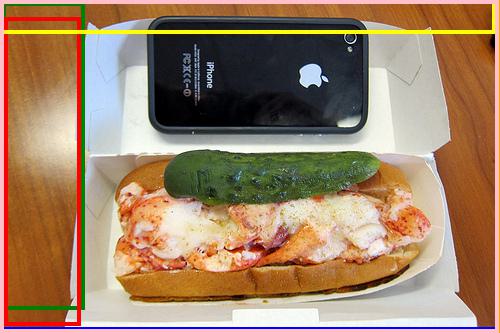}
  \caption{The brown table that is to the left of the black cell phone}
  \label{fig:1c}
\end{subfigure} \hfill
\begin{subfigure}[t]{0.17\linewidth}
  \includegraphics[width=\linewidth,height=2.5cm]{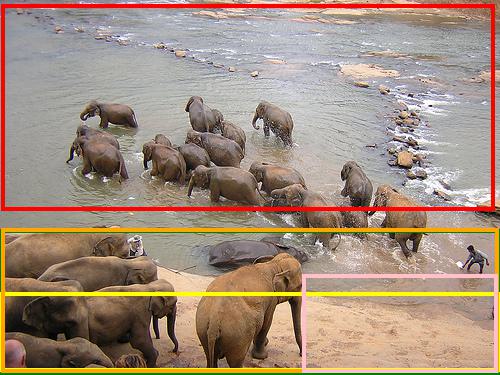}
  \caption{The sandy shore that is near the murky water}
  \label{fig:1d}
\end{subfigure} \hfill
\begin{subfigure}[t]{0.18\linewidth}
  \includegraphics[width=\linewidth,height=2.5cm]{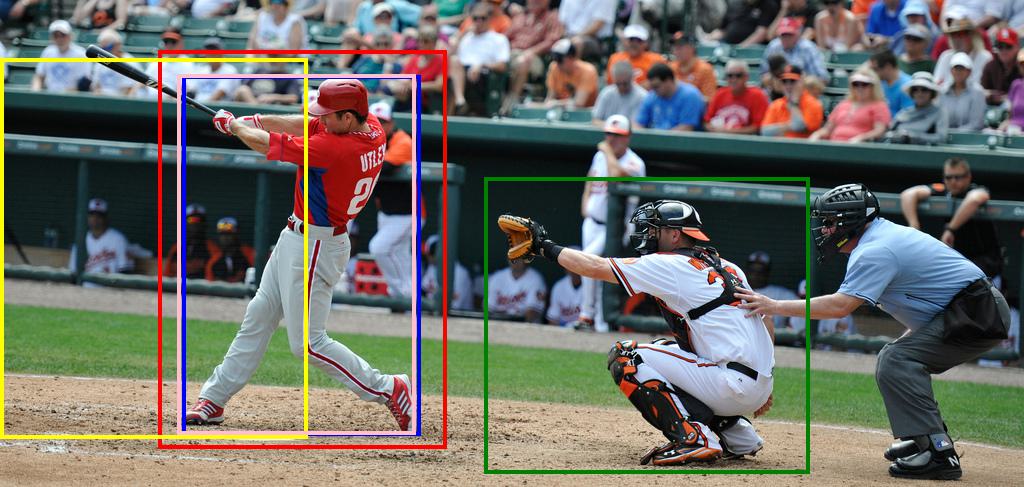}
  \caption{The baseball player that is to the left of the black helmet and to the right of the home plate}
  \label{fig:1e}
\end{subfigure}
\vspace{1em}
\begin{subfigure}[t]{0.21\linewidth}
  \includegraphics[width=\linewidth,height=2.5cm]{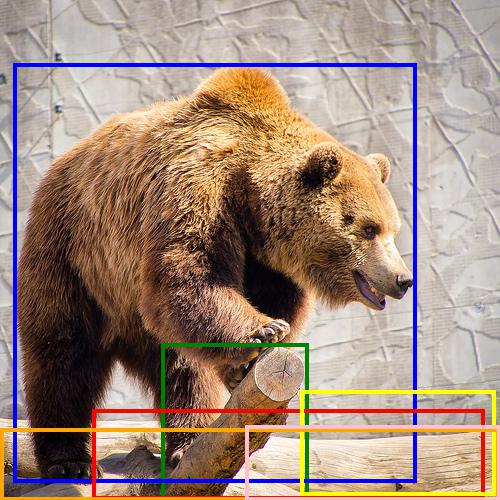}
  \caption{The large branch that is to the right of the log that is behind the large bear}
  \label{fig:1f}
\end{subfigure} \hfill
\begin{subfigure}[t]{0.21\linewidth}
  \includegraphics[width=\linewidth,height=2.5cm]{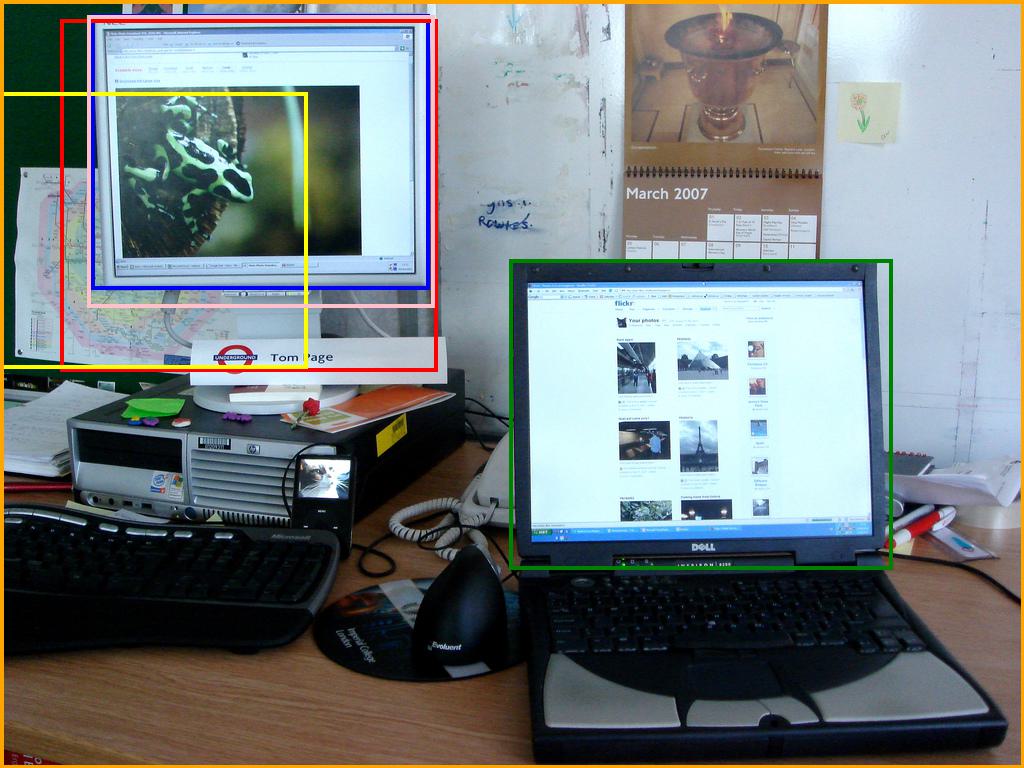}
  \caption{The black monitor that is to the left of the keyboard or on the desk}
  \label{fig:1g}
\end{subfigure} \hfill
\begin{subfigure}[t]{0.21\linewidth}
  \includegraphics[width=\linewidth,height=2.5cm]{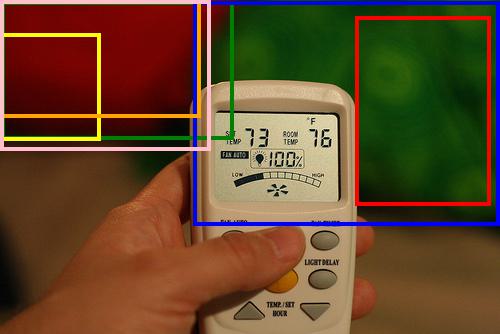}
  \caption{The blanket that is not green and that is not on the bed}
  \label{fig:1h}
\end{subfigure} \hfill
\begin{subfigure}[t]{0.21\linewidth}
  \includegraphics[width=\linewidth,height=2.5cm]{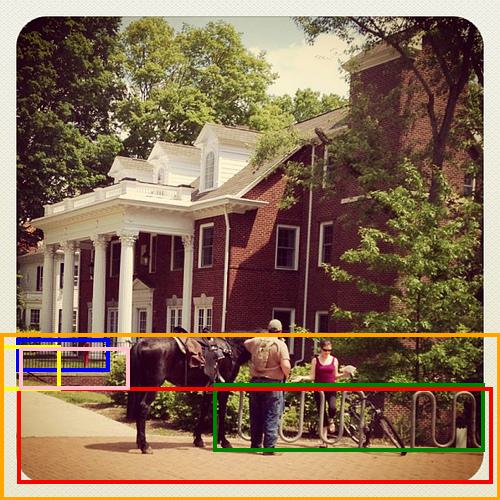}
  \caption{The fence that is not black and that is not to the left of the man}
  \label{fig:1i}
\end{subfigure}
\caption{Figures for qualitative analysis. Bounding box legend - Red: MGA-Net, Blue: GDINO, Yellow: LLaVA, Orange: DeepSeek-VL2, Pink: Qwen2.5-VL, Green: Ground-truth}
\label{fig:example_figures}
\end{figure*}

Some of our important findings are as follows:

\noindent \textbf{(1)} Referring expressions that include spatial relations, in addition to object attributes, result in higher accuracy on the REC task compared to expressions with only attributes. \noindent \textbf{(2)} Increasing the spatial complexity (no. of spatial relations) of an expression affects the performance of the VLMs, but models with explicit compositional learning components maintain the performance. \noindent \textbf{(3)} Expressions involving dynamic spatial relations yield low average accuracy across models, indicating the difficulty in modelling these relations. \noindent \textbf{(4)} The task-specific trained models achieve higher accuracy for expressions with geometric spatial relations (e.g., left of, right of) while the VLMs show relatively better accuracy for expressions having ambiguous relations such as proximity. \noindent \textbf{(5)} The models fail to recognize negated spatial relations in referring expressions in multiple instances, though the extent of this failure varies across models.

\begin{table*}[!ht]
\small
\centering
\begin{tabular}{p{2.2cm} 
                p{1.8cm} 
                p{1.8cm} 
                p{2.3cm} 
                p{2.9cm} 
                p{2.2cm}}
\toprule
\textbf{Dataset} & \textbf{Object Categories} & \textbf{Image Source} & \textbf{Avg. Length of Expression} & \textbf{Avg. No. of Objects per Image} & \textbf{No. of Spatial Relations} \\
\midrule
RefCOCO     & 80  & Real-world & 3.6  & 10.6 & 59 \\
RefCOCOg    & 80  & Real-world & 8.4  & 8.2  & 72 \\
CLEVR-Ref+  & 3   & Synthetic  & 22.4 & 6.5  & 4 \\
CopsRef     & 508 & Real-world & 14.4 & 17.4 & 51 \\
\bottomrule
\end{tabular}
\caption{Statistics of Popular Referring Expression Comprehension Datasets. For the last column, the relation types are taken from various resources explained in Section~\ref{sec:rel work} and~\cite{marchi}, in addition to the relations in Table~\ref{table: our work cat}.}
\label{table: data-stats}
\end{table*}

\begin{table*}[!ht]
\small
\begin{tabular}{p{1.7cm} p{1.1cm} p{12cm}}
\toprule
\textbf{Category} & \textbf{Number} & \textbf{Spatial Relations} \\
\midrule
Absolute & 56 & on the right, on the left, in the middle, in the center, from the right, from the left\\
Adjacency & 14 & attached, against, on the side, on the back, on the front, on the edge \\
Directional & 29 & falling off, along, through, across, down, up, hanging from, coming from, around\\
Orientation & 0 & facing \\
Projective & 2361 & on top of, beneath, beside, behind, to the left, to the right, under, above, in front of, over, below, underneath \\ 
Proximity & 217 & by, close to, near \\ 
Topological & 1054 & connected, contain, with, surrounding, surrounded by, inside, between, touching, out of, at, in, on \\
Unallocated & 56 & next to, enclosing \\
\bottomrule
\end{tabular}
\caption{\label{citation-guide}
Category-wise relation split and number of referring expressions in the CopsRef (\citealp{copsref}) test set with 1 spatial relation in each category. While there are no referring expressions with only one relation from the orientation category, these relations co-occur with relations from other categories in some expressions.
}
\label{table: our work cat}
\end{table*}

\section{Related Work}\label{sec:rel work}
Previous works have conducted a broad analysis on the ability of VLMs to perform multimodal perception and reasoning tasks, such as Spatial Reasoning, Multimodal conversation, etc. Many comprehensive real-world benchmarks have been introduced to test multiple VLM capabilities. (\citealp{llava,cb300k,mmbench,blink}).

Some works (\citealp{subramaniamreclip,roschpi}) focus solely on spatial analysis of VLMs. \citealp{pic1k} go a step further to analyze the role of each modality in spatial reasoning. However, these works do not analyze the factors that affect the spatial reasoning ability of the VLMs. Another class of works performs a category-wise analysis of spatial relations, either based on their spatial properties (\citealp{vsr, sr2d}) or their linguistic properties and complexity (\citealp{film}). In contrast, \citealp{whatsup} analyze the effects of spatial biases in the datasets for the REC task performance.

\noindent \paragraph{Task Complexity and Interpretability.} The above-mentioned works use image-caption agreement as their evaluation task. Due to the inherent limitations of this task, these works simplified the expressions to have only 2 objects and 1 spatial relation. To improve the interpretability of model output, synthetic datasets have been used instead of real-world images (\citealp{subramaniamreclip,clipcdsm}). However, it simplifies the problem due to bounded expressivity (a limited set of attributes and spatial relations). On the other hand, REC models output bounding boxes around the target objects. Analyzing the characteristics of these boxes helps identify the parts of the input that the models fail to process. This enables comparative analysis of expressions with 0, 1, or more spatial relations, a unique feature of our work. The REC task also enables us to test the models over images of different visual complexities (single or multiple instances of objects in an image).

\noindent \paragraph{Related Tasks - Embodied Settings.} Beyond the REC task, spatial reasoning work also spans embodied settings. Spatial affordance prediction involves localizing precise action points in the image based on a spatial language instruction\cite{robopoint}. Benchmarks such as Room-to-Room (R2R) \cite {R2R} involve agents following natural-language route descriptions through real buildings, requiring robust spatial reasoning over landmarks and paths. Embodied Spatial Analysis, which focuses on the effects of different perspectives and non-verbal cues on the spatial reasoning capabilities of VLMs \cite{islam1,islam2}.

\section{Dataset}\label{sec:dataset}
Table~\ref{table: data-stats} shows the key characteristics of some popular REC datasets. We chose CopsRef (\citealp{copsref}) because among real-world datasets, it has the longest referring expressions, which go beyond describing the simple, distinctive properties of the objects. CopsRef is also a highly spatial dataset, as 90\% of expressions consist of spatial relations. Examples of such referring expressions and the corresponding images are given in Figure~\ref{fig:example_figures}. Table~\ref{table: our work cat} shows the category-wise split of the 51 spatial relations we identified in the CopsRef test dataset. For explanations of each category, refer to Appendix~\ref{app: cat ex}.

\begin{table}[!ht]
\small
\centering
\begin{tabular}{p{1.8cm} p{2.2cm} p{1.5cm}}
\toprule
\textbf{Category} & \textbf{No. of relations} & \textbf{No. of expressions} \\
\midrule
None & 0 & 1202 \\
One & 1 & 3787 \\
Two-chained & 2 & 1324 \\
Two-and & 2 & 3890 \\
Two-or & 2 & 2203 \\
Three & 3 & 180 \\
\bottomrule
\end{tabular}
\caption{\label{table: Cat-wise}
Frequency of occurrence of spatial relations in referring expressions.
}
\end{table}

Table~\ref{table: Cat-wise} shows the number of expressions having 0, 1, 2, and 3 spatial relations. For expressions with 2 spatial relations, we have introduced three categories. The first category, ‘Two-chained’, includes expressions where the spatial clauses are chained sequentially. The second category, ‘Two-and’, contains expressions where the referred object satisfies both spatial clauses. Finally, ‘Two-or’ consists of expressions where the referred object satisfies either of the two spatial clauses. Figures~\ref{fig:1e}-\ref{fig:1g} illustrate examples of the three categories.

\section{Approach}
In our analysis, we consider spatial reasoning to include both basic grounding (locating the right entities) and spatial relations (reasoning about where entities are relative to one another). These aspects are inseparable in real scenes as attributes and object properties (size, shape, rigidity, affordances) determine which relations are even plausible. Eg: a rock submerges in water, while a feather tends to float on the surface of water. Decoupling them might yield an unrealistic toy setup. We use REC as the analysis medium because it exposes failures in either component. We therefore seek to the following research questions:\\
\textbf{RQ1.} Which spatial relation categories result in low accuracy for REC models? \textbf{RQ2.} How do different model characteristics/architectures influence the REC task accuracy for certain spatial relation categories compared to the others? \textbf{RQ3.} Does the inclusion of spatial relations increase or decrease the accuracy of REC models? \textbf{RQ4.} How does the number of spatial relations in the expressions affect the accuracy across different types of models? \textbf{RQ5.} Do the REC models accurately recognize negated spatial relations in expressions?

To answer these questions, we outline our research methodology and the designed experiments in this section:
\subsection{Models Description}\label{subsec: res met - models}
\noindent\paragraph{MGA-Net. (\citealp{mganet})} It is an REC task-specific model with a compositional architecture designed to handle complex expressions. The model uses soft attention to decompose the expression and builds a relational graph among objects using the connecting spatial relations and attributes. A Gated Graph Neural Network then performs multi-step reasoning over this graph. We use Faster R-CNN (\citealp{frcnn}) to generate object proposals and extract their features using a pre-trained ResNet-101. The model is designed to use the individual bounding box features instead of the image as a whole.
\noindent\paragraph{Grounding DINO (GDINO). (\citealp{gdino})} It is an open-set object detector VLM with both vision and language backbones, whose outputs are fused at multiple levels. Grounded pre-training with contrastive loss makes it well-suited for the REC task. We use the Swin-B vision backbone and the CLIP-text encoder (\citealp{clip}) for the language backbone. We filter all bounding box detections for an expression using their output labels to see which detections match the target entity. Then we select the detection with the highest confidence score.
\noindent\paragraph{LLaVA. (\citealp{llava})} It is a general-purpose VLM that connects an open-set vision encoder from CLIP with a language decoder. The model is trained end-to-end, which involves general visual instruction tuning for aligning the vision and language modalities.
\noindent\paragraph{DeepSeek-VL2. (\citealp{deepseek})} It is a popular vision-language model with a dynamic tiling-based vision encoder (SigLIP) and a Mixture-of-Experts (MoE) language model, whose features are aligned using a vision-language adaptor. The model has been trained for grounded conversation (return the bounding box for an expression) using popular REC datasets like RefCOCO, which feature simple expressions with limited relational complexity. We employ the DeepSeek-VL2 Tiny variant of the model to suit the available computational resouces.
\noindent\paragraph{Qwen2.5-VL. (\citealp{qwen})} It is a multimodal model built over a Vision Transformer (ViT) encoder having native-resolution support to preserve spatial information and a Qwen2.5 LLM decoder. An MLP-based vision-language merger aligns these two modalities. Qwen2.5-VL is supervised using both point and bounding box annotations for grounding tasks. The model is trained on diverse referring expressions with over 10,000 real and synthetic object categories.
\noindent\paragraph{OWL-ViT. (\citealp{owlvit})} It is an object detector baseline that only takes the target object's label as the input instead of the referring expression. It is an open-set object detector, suitable for the CopsRef dataset because its expressions include entities from the Visual Genome (\citealp{visualgenome}) Scene Graphs, some of which are absent in datasets used to train famous closed-set detectors like YOLO (\citealp{yolo}). It also has a simple architecture with a Vision transformer and CLIP for zero-shot image–label alignment, making it an ideal baseline.

\subsection{Evaluation Dataset Splits}\label{subsubsec: setting}
We create the following dataset test splits for evaluation and answering the earlier mentioned research questions, RQ1-RQ5. 
\noindent\paragraph{Fine-grained Spatial Relations Split.} In the test dataset, we split the expressions with 1 spatial relation using the categories shown in Table~\ref{table: our work cat}. Using the categories from Table~\ref{table: Cat-wise}, we split the remaining expressions based on the number of spatial relations they contain. Then, we rank the models based on their accuracy for each category. To statistically find the correlation between the models' performance across categories, we employ the Kendall Tau Independence Test. The details about the test can be found in Appendix~\ref{app: kendall}.
\noindent\paragraph{Visual Complexity Split.} To observe the effect of visual complexity on model performance, we split the test dataset into two parts. The first part has images that have multiple instances of one or more objects mentioned in the referring expressions. The second part has images with at most one instance of every object mentioned in the expression. We perform this splitting by first collecting the entities in each expression using spaCy\footnote{\url{https://spacy.io/}} and then employing GDINO to find the number of instances in the image for each of the collected entities.
\begin{table}[!ht]
\small
\centering
\begin{tabular}{p{3.5cm} p{3cm}}
\toprule
\textbf{Model} & \textbf{Accuracy (\%)} \\
\midrule
MGA-Net & $62.92 \pm 0.11$ \\
GDINO & $70.93 \pm 0.01$ \\
LLaVA & $34.96 \pm 0.03$ \\
DeepSeek-VL2 & $30.07 \pm 0$ \\
Qwen2.5-VL & $67.00 \pm 0$ \\
OWL-ViT & $56.34 \pm 0$ \\
\bottomrule
\end{tabular}
\caption{
Comprehension Accuracies}
\label{table: Comp-acc}
\end{table}
\begin{table*}[!htb]
\small
\centering
\begin{tabular}{p{1.7cm}
                | p{1.7cm} p{0.3cm}
                | p{1.7cm} p{0.3cm}
                | p{1.7cm} p{0.3cm}
                | p{1.5cm} p{0.3cm}
                | p{1.5cm} p{0.3cm}}
\toprule
\textbf{Category} &
\multicolumn{2}{c|}{\textbf{MGA-Net}} &
\multicolumn{2}{c|}{\textbf{GDINO}} &
\multicolumn{2}{c|}{\textbf{LLaVA}} &
\multicolumn{2}{c|}{\textbf{DeepSeek-VL2}} &
\multicolumn{2}{c}{\textbf{Qwen2.5-VL}} \\
& Acc (\%) & \makebox[0.4cm][r]{Rank}
& Acc (\%) & \makebox[0.4cm][r]{Rank}
& Acc (\%) & \makebox[0.4cm][r]{Rank}
& Acc (\%) & \makebox[0.4cm][r]{Rank}
& Acc (\%) & \makebox[0.4cm][r]{Rank} \\
\midrule
Absolute       & $70.24 \pm 2.22$ & 1  & $82.14 \pm 0$    & 2  & $44.64 \pm 0$    & 4  & $47.37 \pm 0$    & 3  & $80.70 \pm 0$     & 1  \\
Adjacency      & $52.38 \pm 3.37$ & 12 & $78.57 \pm 0$    & 4  & $50.00 \pm 0$       & 1  & $71.43 \pm 0$    & 1  & $71.43 \pm 0$    & 4  \\
Directional    & $52.87 \pm 3.25$ & 11 & $65.52 \pm 0$    & 12 & $27.59 \pm 0$    & 12 & $41.40 \pm 0$    & 5  & $70.69 \pm 0$    & 5  \\
Projective     & $64.07 \pm 0.08$ & 4  & $69.12 \pm 0$    & 8  & $36.19 \pm 0.08$ & 6  & $34.57 \pm 0$    & 7  & $70.52 \pm 0$    & 6  \\
Proximity      & $62.83 \pm 0.22$ & 8  & $80.65 \pm 0$    & 3  & $46.84 \pm 0.22$ & 3  & $45.16 \pm 0$    & 4  & $71.89 \pm 0$    & 3  \\
Topological    & $67.32 \pm 0.49$ & 2  & $83.02 \pm 0$    & 1  & $48.51 \pm 0.09$ & 2  & $59.87 \pm 0$    & 2  & $72.91 \pm 0$    & 2  \\
Unallocated    & $63.09 \pm 0.84$ & 6  & $75.00 \pm 0$       & 5  & $35.71 \pm 0$    & 7  & $30.36 \pm 0$    & 9  & $68.75 \pm 0$    & 7  \\
None           & $62.39 \pm 0.58$ & 9  & $73.88 \pm 0$    & 6  & $42.89 \pm 0.17$ & 5  & $41.10 \pm 0$    & 6  & $66.14 \pm 0$    & 8  \\
Two-chained    & $63.21 \pm 0.22$ & 5  & $70.67 \pm 0.03$ & 7  & $30.82 \pm 0$    & 9  & $18.13 \pm 0$    & 10 & $65.78 \pm 0$    & 9  \\
Two-and        & $62.98 \pm 0.09$ & 7  & $68.45 \pm 0.01$ & 10 & $31.57 \pm 0.07$ & 8  & $31.45 \pm 0$    & 8  & $65.69 \pm 0$    & 10 \\
Two-or         & $59.39 \pm 0.21$ & 10 & $68.97 \pm 0.01$ & 9  & $30.26 \pm 0.04$ & 10 & $12.68 \pm 0$    & 11 & $63.20 \pm 0$     & 11 \\
Three          & $65.18 \pm 2.1$  & 3  & $67.78 \pm 0$    & 11 & $30.19 \pm 0.26$ & 11 & $11.11 \pm 0$    & 12 & $62.78 \pm 0$    & 12 \\
\bottomrule
\end{tabular}
\caption{Category-wise accuracy and ranking}
\label{table: Cat-wise res}
\end{table*}
\noindent\paragraph{Negation Analysis Split.} In our analysis, we found that models have difficulties in grounding spatial expressions with negations. Therefore, we created a test split for a more accurate evaluation and a deeper analysis of negated spatial expressions. We collected expressions that include the keyword ‘not’ and divided them into two sets according to the number of occurring negations (1 or 2). Then, we collected those expressions for which all the models give an IoU of less than 0.5. For each expression, we perform a qualitative analysis to verify whether the errors are due to misinterpreting the negations or conflation of other errors. We limit our analysis to the results from the first run of the models to facilitate the instance-wise analysis.

\section{Results}
\textbf{Hardware.} For inference of GDINO, LLaVA, OWLViT, and training of MGA-Net, we use the Quadro RTX 6000. Due to the heavy computational (GPU) requirements of DeepSeek-VL2 Tiny and Qwen2.5-VL, we use NVIDIA A100-SXM4 for their inference.\\
\textbf{Evaluation.} We evaluate the models using the Intersection over Union (IoU) metric. Following previous works (\citealp{mattnet,recnew}), we consider the output as a correct comprehension if the IoU is greater than 0.5. We calculate the accuracy as the fraction of data points that have an IoU \textgreater 0.5. We run each model three times (both training and testing for MGA-Net, and inference for the other models) to ensure the statistical significance of the evaluation.
\subsection{Evaluation on Referring Expressions}
We report the average accuracy and standard deviation across the three runs for each model. Since we retrain MGA-Net for each run, the model predictions slightly change every time, resulting in the highest standard deviation among all models. VLMs and the baseline have zero or near-zero standard deviation since we test them zero-shot. This also follows for the future result tables.

GDINO performs the best, followed by Qwen and MGA-Net. LLaVA and DeepSeek perform worse than the baseline. A possible reason is that they don't possess a bounding box regression architecture. Another reason could be that while LLaVA doesn't have visual grounding instructions during pre-training, DeepSeek's training referring expressions often lack spatial complexity and use simple, non-relational phrases.
\begin{table}[h!]
\small
\centering
\begin{tabular}{p{1.2cm} p{0.7cm} p{0.8cm} p{0.8cm} p{0.8cm} p{0.8cm}}
\toprule
 & \textbf{MGA-Net} & \textbf{GDINO} & \textbf{LLaVA} & \textbf{Deep-Seek} & \textbf{Qwen} \\
\midrule
MGA-Net   & 1.00   & 0.18 & 0.09 & -0.12 & 0.11 \\
GDINO     & 0.18   & 1.00 & 0.73 & 0.52  & 0.63 \\
LLaVA     & 0.09   & 0.73 & 1.00 & 0.73  & 0.53 \\
DeepSeek  & -0.12  & 0.52 & 0.73 & 1.00  & 0.75 \\
Qwen      & 0.11   & 0.63 & 0.53 & 0.75  & 1.00  \\
\bottomrule
\end{tabular}
\caption{Kendall Tau Independence Test results for category-wise ranks. DeepSeek model's version is VL2 and Qwen model's version is 2.5-VL.}
\label{table: Kendall tau cat}
\end{table}
\begin{table*}[!htb]
\small
\centering
\begin{tabular}{lrrrrrr}
\toprule
\textbf{No. of relations} & \textbf{MGA-Net} & \textbf{GDINO} & \textbf{LLaVA} & \textbf{DeepSeek} & \textbf{OWL-ViT} & \textbf{Qwen} \\
\midrule
None   & $62.39 \pm 0.59$ & $73.88 \pm 0$    & $42.89 \pm 0.17$ & $41.10 \pm 0$ & $67.80 \pm 0$  & $66.14 \pm 0$ \\
One    & $64.85 \pm 0.13$ & $73.94 \pm 0$    & $40.33 \pm 0.01$ & $42.53 \pm 0$ & $60.50 \pm 0$  & $71.40 \pm 0$  \\
Two    & $61.96 \pm 0.07$ & $69.00 \pm 0$       & $31.05 \pm 0.06$ & $23.50 \pm 0$ & $52.39 \pm 0$ & $64.96 \pm 0$ \\
Three  & $65.18 \pm 2.1$  & $67.78 \pm 0$    & $30.19 \pm 0.26$ & $11.11 \pm 0$ & $55.00 \pm 0$    & $62.78 \pm 0$ \\
\bottomrule
\end{tabular}
\caption{Spatial relation frequency results and ranking}
\label{table: Freq-wise}
\end{table*}
\subsection{Evaluation on Fine-grained Relations}\label{subsec: Eval fgsp}
Table~\ref{table: Cat-wise res} shows a few general trends in results. The top 3-4 categories which each model performs the best for are categories with a single spatial relation. Among those, all the models perform well for the Topological and Absolute categories.

To answer \textbf{RQ1}, we observed that among all categories with a single spatial relation, the average accuracy across models is lowest for the Directional category expressions. A possible reason is that the spatial configurations of the involved objects are dynamic, as they vary from image to image for the same spatial relation. This makes it difficult for the models to learn common patterns for recognizing these relations, resulting in low accuracy.
\subsection{Impact of Multiple Spatial Relations}\label{subsec: results - model-wise-rel}
Table~\ref{table: Kendall tau cat} shows the Kendall Tau correlation values for all pairs of models. From the Kendall Tau Independence test, we observed that while all the VLMs are correlated, MGA-Net is not correlated with any of them. We study the reasons behind MGA-Net and VLMs differing in category-wise performance.

Among spatial categories of MGA-Net and VLMs, the major difference occurs with the Proximity and Projective categories. To answer \textbf{RQ2}, we can observe that the ‘Proximity’ category ranked similarly for the VLMs and higher than MGA-Net. On the other hand, ‘Projective’ has a higher rank for MGA-Net compared to all VLMs. We can see that MGA-Net prefers geometric spatial relations like left of, on top of, etc., as it explicitly encodes the relative locations of bounding boxes, which helps represent such relations. On the other hand, the VLMs have a better ranking for ambiguous relation categories that do not specify a clear distance or geometric direction (e.g., by, close to). This is because the vision backbones of the VLMs utilize the entire image and help capture relations between a region in the image and its surrounding regions, unlike MGA-Net, which only receives the detected bounding boxes as input.

Table~\ref{table: Freq-wise} shows the performance of all models and the OWL-ViT baseline for expressions having different numbers of spatial relations. We observe that VLMs perform considerably better for expressions with 0/1 spatial relations compared to expressions with 2/3 spatial relations.  This proves that VLMs find it comparatively difficult to ground multiple spatial relations. Among the VLMs, GDINO has the least drop in accuracy as no. of relations increase, while DeepSeek has the maximum drop. However, MGA-Net utilizes its compositional learning architecture to handle multi-step reasoning, resulting in a similar performance for all categories.

An interesting observation is that the performance of the baseline considerably drops for the ‘Two' and ‘Three' categories, even though the spatial relations aren't being passed as input to the baseline. The reason might be that 41.4\% of these images have multiple instances of objects, the impact of which is explained in the next section.

Now, to answer \textbf{RQ3}, we observe in Table~\ref{table: Freq-wise} that for all models except GDINO and LLaVA, the performance is better for expressions with one spatial relation than no spatial relations. Table~\ref{table: Cat-wise res} further shows that among the seven categories of single spatial relations, GDINO and LLaVa perform better for 4-5 of them compared to expressions with no relations. Thus, we can conclude that in a setup involving visual and linguistic ambiguity (such as ours), spatial relations along with visual attributes often aid the models in grounding the expressions, compared to the attributes alone. This is also reinforced by the results of the baseline. From Table~\ref{table: Freq-wise}, we can observe that while the baseline gives the second-best performance for expressions with no spatial relations, it drops to the third place for expressions with one spatial relation, with a 7.3\% reduction in performance. This is because the baseline doesn't have access to the spatial relations.

Finally, Table~\ref{table: Freq-wise} helps us answer \textbf{RQ4} as it shows the effect of increasing spatial relations on the performance of MGA-Net versus the VLMs (as discussed before).

\subsection{Impact of Visual Complexity}
\begin{table}[!ht]
\small
\centering
\begin{tabular}{@{}l*{3}{l}@{}}
\toprule
\textbf{Model}    & \textbf{Acc Single} & \textbf{Acc Multi}\\ \midrule
MGA-Net   & $64.91 \pm 0.15$           & $59.61 \pm 0.04$   \\
G-DINO    & $72.54 \pm 0.01$           & $68.94 \pm 0.01$   \\
LLaVA     & $37.69 \pm 0.01$           & $30.43 \pm 0.1$    \\
DeepSeek-VL2     & $32.53 \pm 0$           & $27.46 \pm 0$    \\
Qwen2.5-VL     & $68.47 \pm 0$           & $63.76 \pm 0$    \\
OWL-ViT  & $59.71 \pm 0$              & $51.30 \pm 0$       \\
\bottomrule
\end{tabular}
\caption{Results for accuracy in different visual complexity settings. All accuracies are in (\%).}
\label{tab:multi}
\end{table}
Out of 12586 test instances, we found that 4730 instances have images with multiple instances of objects mentioned in the referring expressions. Table~\ref{tab:multi} shows the accuracies of all models and the OWL-ViT baseline for images with a single instance (‘Acc Single' column) and multiple instances (‘Acc Multi' column). The models perform better for the single instance images by 5.7\% on average compared to the multi-instance images. The 8.4\% performance drop of the baseline for multi-instance images proves that the images are indeed complex and require more than just the label as the input for grounding the right object. Excluding the baseline, LLaVa has the steepest performance drop, showing that grounded pre-training also plays a crucial role in helping the models ground the right object instance in multi-instance images.

\subsection{Impact of Negation}
\begin{table}[!ht]
\small
\centering
\begin{tabular}{@{}l*{2}{l}@{}}
\toprule
                 & \textbf{2 Negations}          & \textbf{1 Negation} \\ \midrule
Negation failure    & 92.86                     &  93.75                 \\
DeepSeek        & 88.1                   & 87.5                       \\
GDINO            & 78.57                     & 62.5                    \\
LLaVA            & 35.71                     & 25                    \\
MGA-Net          & 52.38                     & 50                    \\
Qwen2.5-VL       & 50                         & 62.5                 \\
\bottomrule
\end{tabular}
\caption{Results for negation analysis. All values are in (\%) w.r.t the total failure instances (42 for 2 Negations and 16 for 1 Negation).}
\label{tab:negations}
\end{table}
We obtained 16 expressions with 1 ‘not’ and 42 expressions with 2 ‘not’s for which all models gave incorrect predictions (as shown in Table \ref{tab:negations}). The ‘Negation failure' row gives the percentage of instances for which at least 1 model gives an incorrect prediction due to failing to recognize negations and not due to conflation of other errors. We can observe that DeepSeek has the highest percentage of failure instances, while GDINO records the second most failures. However, while DeepSeek has the worst comprehension accuracy for the REC task, GDINO has the highest accuracy (refer to Table \ref{table: Comp-acc}).

A possible reason for VLMs like LLaVA and Qwen2.5-VL performing better than GDINO in recognizing negations is due to their superior language backbones (Vicuna (\citealp{vicuna}) and Qwen-2.5 LLM, respectively) that have better language understanding (including negations) compared to GDINO's CLIP text encoder. MGA-Net outperforms GDINO since its training involves expressions with negations, increasing its ability to comprehend negations during testing. Hence, to answer \textbf{RQ5}, we observe that while all REC models face issues with recognizing negations, certain model characteristics and training paradigms might reduce the failure cases.

\begin{table}[!ht]
\small
\centering
  \begin{tabular}{@{}l*{4}{l}|@{}}
    \toprule
    \textbf{Models} & \textbf{Negations} & \textbf{Precision} & \textbf{Recall} \\
    \midrule
    MGA-Net & 1 & 53.60 & 70.8 \\
    MGA-Net & 2 & 41.38 & 51 \\
    LLaVA & 1 & 64.54 & 47.23 \\
    LLaVA & 2 & 60.35 & 41 \\
    \bottomrule
  \end{tabular}
\caption{Negation Precision (\%) and Recall (\%): MGA-Net vs. LLaVA}
\label{table: pre-rec neg}
\end{table}

Another interesting observation was for the outputs of MGA-Net and LLaVA models when they are close to the target object. From Table~\ref{table: pre-rec neg}, we can see that while LLaVA has a better precision in such cases, MGA-Net has a better recall.
\section{Qualitative Analysis}
In this section, we provide a qualitative analysis of certain issues faced by the models in handling referring expressions.
\subsection{Directional Relations}\label{subsec: qual analysis - Directional}
The expressions pertaining to Figures~\ref{fig:1a} and \ref{fig:1b} consist of the same spatial relation (‘around’). In the first figure, the wrapping of the napkin around the hot dog only makes the napkin partially visible. But in the second figure, the white box around the mirror is almost entirely visible. This shows how the interpretation of ‘around’ is highly dependent on the configuration of the involved objects. For the first image, LLaVA fails to precisely localize the object, while MGA-Net only returns a part of the napkin that is visible. In the second image, both models fail to localize the object. DeepSeek fails to output a bounding box for the first image but gets it right for the second.
\subsection{Projective and Proximity Relations}\label{subsec: qual analysis - P&P}
Figure~\ref{fig:1c} shows an example of a Projective relation (‘to the left’). MGA-Net succeeds in returning the correct part of the table that is to the left of the phone. While GDINO, DeepSeek, and Qwen simply return the entire table, LLaVA identifies the wrong part. This shows the ability of MGA-Net to comprehend projective relations better, particularly when the target object is not apparent. An example of Proximity relations is in Figure~\ref{fig:1d}, where LLaVA, GDINO, and DeepSeek return the shore that is ‘near’ the murky water, but MGA-Net fails to do so. Interestingly, Qwen only returns that part of the shore which isn't occluded by the elephants.
\subsection{Multiple Spatial Relations}
For ‘Two-and' category expressions, the models sometimes only satisfy one of the spatial clauses. This often happens if multiple objects of the same class are in the image. For example, in Figure~\ref{fig:1e}'s prediction for all models (except DeepSeek, which doesn't return any output), the output baseball player is to the left of the black helmet but is not to the right of the home plate.

Similarly, for 'Two-chained' category expressions, the models sometimes do not consider the entire expression. For example, in Figure~\ref{fig:1f}, MGA-Net, LLaVA, DeepSeek, and Qwen return the ‘log that is behind the large bear’, and GDINO returns the bear itself. None of the models consider the ‘large branch’ part of the expression, which should have been the output.

Finally, for 'Two-or' category expressions, the models might consider only one spatial clause. Consequently, it returns an object satisfying that clause but not the additional attributes mentioned in the expression. For example, in Figure~\ref{fig:1g}, the models return the monitor that is to the ‘left of the keyboard’, but it does not satisfy the color attribute.
\subsection{Negation}
Figures \ref{fig:1h} and \ref{fig:1i} show two cases where all models fail to recognize negation. In \ref{fig:1h}, we can observe that while MGA-Net is wrong, LLaVA is close to the ground truth but partially covers the target object (high precision, low recall). In \ref{fig:1i}, while LLaVA, GDINO and Qwen are wrong, MGA-Net is closest to the ground truth but covers an excess area (low precision, high recall).

\section{Conclusion}
Spatial reasoning is an integral aspect of cognitive reasoning and embodied AI tasks. However, recent studies have shown that state-of-the-art VLMs often fail to accurately comprehend spatial relations. To analyze the limitations of these models, we evaluate their spatial understanding using the referring expression comprehension task. We picked multiple models, including Vision-language models (LLaVA, GDINO, DeepSeek, Qwen) and task-specific models (MGA-Net). We observed that the VLMs that are trained in the wild with visual and textual data perform worse in grounding. The models perform the worst in grounding Directional relations on average. However, the VLMs do better in vague relations such as proximity, while the task-specific models are better in geometrically well-defined relations such as left and right. While using spatial relations increases the grounding accuracy, using multiple relations makes the reasoning more challenging for all models, with a higher impact on VLMs. However, MGA-Net maintains its performance for complex spatial expressions due to its compositional learning architecture. In the presence of visual complexity, the performance of all models drops, but DeepSeek and LLaVa's performances are affected the most due to a lack of grounded pre-training with complex expressions. Finally, both VLMs and task-specific models have failure cases when grounding expressions that include negation. These findings shed light on the gaps for future work on Vision-language models.

\section{Future Directions}
While we consider a comprehensive set of spatial relations for spatial analysis, these relations are atomic and don't include metric properties such as distances. We plan to extend our analysis to include metric relations in the future.

From this work, an important inference is that while increasing the number of parameters in VLMs can improve performance on expressions with simple spatial relations, architectural changes are needed to handle novel, complex compositions effectively. Unlike VLMs, MGA-Net maintains consistent performance across spatial complexities by using a soft attention module that decomposes expressions into semantic components for compositional reasoning. This suggests expression decomposition can enhance VLM generalization. Alternative strategies (\citealp{compsurvey}) could be using multi-modal transformer models (\citealp{sikarwar2022can, qiu2021systematic}) and techniques such as weight sharing across transformer layers or ‘Pushdown layers’ with recursive language understanding (\citealp{murty2023pushdown}). Another promising direction is Neuro-symbolic processing (\citealp{kamali2024nesycoco,hsu2024}), which involves generating symbolic programs from expressions using LLMs and conducting explicit symbolic compositions before grounding into visual modality. We plan to explore integrating such techniques with VLMs.

Another issue to address is the VLMs' inability to comprehend negations. Our experiments with the VLMs and MGA-Net suggest that augmenting the training/instruction tuning with synthetically generated negated expressions can help. Additionally, we also plan to formulate contrastive learning objectives to penalize the model when it fails to comprehend negations.

\section*{Limitation}
This paper is an analysis study on the shortcomings of the vision and language models when it comes to fine-grained spatial reasoning.  Our analysis covers a variety of vision and language models including closed and open ones. However, the number of language models that we cover is by no means exhaustive. Spatial reasoning is important for many downstream applications however, we chose referring expressions as a platform that can demonstrate the challenges in both language and vision sides. While spatial understanding becomes a very important skill for embodied AI, in this work we do not consider the interaction with the environment and we do not consider the change of perspective. Our study can serve as a complement studies in this area that can provide insight into the difficulties of spatial language understanding and grounding language into visual perception. Our study was constrained by the cost of proprietary LLMs and the computational resources for open source ones. 
\bibliography{custom}

\appendix

\section{Description of spatial categories}\label{app: cat ex} 
For our analysis, we utilize the spatial categories introduced by  (\citealp{marchi}) and replace the 'Cardinal Direction' category with 'Absolute'. The descriptions and examples for the chosen categories are as follows:
\begin{enumerate}
    \item \textbf{Absolute}: Consists of relations that describe the location of an object in an absolute manner and not in relation to another object. \\
    E.g.: man on the right that is standing and wearing gray pant 
    \item \textbf{Adjacency}: Consists of relations that describe the close, side-by-side positioning of two objects. They may or may not imply a particular direction. \\
    E.g.: The large poster that is leaning against the wall
    \item \textbf{Directional}: Consists of dynamic action verbs / directional relations. They describe the movement or change in position of an object relative to other objects in the image. The interpretation of these relations heavily relies on the configuration of the involved objects and/or the dynamic spatial relationship between them. \\
    E.g.: The gray car that is driving down the road
    \item \textbf{Orientation}: Consists of relations which describe the orientation of an object w.r.t another object. \\
    E.g.: The sitting dog that is facing the window that is to the right of the mirror
    \item \textbf{Projective}: Consists of relations that indicate the concrete spatial relationship between two objects, i.e., these relations can be quantified in terms of the coordinates of the two objects. \\
    E.g.: The black oven that is above the drawer
    \item \textbf{Proximity}: Consists of relations that indicate that two objects are near each other without giving a specific directional relationship. \\
    E.g.: The blue chair that is close to the white monitor
    \item \textbf{Topological}: Consists of relations that indicate the broader arrangement or the containment of an object w.r.t another object \\
    E.g.: The silver train that is at the colorful station
    \item \textbf{Unallocated}: Consists of relations that cannot be allocated to any of the above categories.
\end{enumerate}

\section{Kendall Tau Independence Test}\label{app: kendall}
To compare the models' performances across the categories, we employ a statistical test known as the Kendall Tau Independence Test.
It evaluates the degree of similarity between two sets of ranks given to the same set of objects. We calculate the Kendall rank coefficient (\begin{math}\tau\end{math}), which yields the correlation between two ranked lists. Given \begin{math}\tau\end{math} value, we calculate the $z$ statistic, which follows standard normal distribution, as:
\begin{equation}
    z = 3*\tau*\sqrt{n(n-1)}/\sqrt{2(2n+5)}.
    \label{eq:kendall_z}
\end{equation}
Using the 2-tailed p-test at 0.05 level of significance, we test the following:
\begin{itemize}
\item \textbf{Null hypothesis}: There is no correlation between the two ranked lists.
\item \textbf{Alternative hypothesis}: There is a correlation between the two ranked 
\end{itemize}

\section{Additional Model Settings and Experiments}
\begin{table}[!ht]
\small
\centering
\begin{tabular}{p{4cm} p{2.5cm}}
\toprule
\textbf{Model} & \textbf{Accuracy (\%)} \\
\midrule
MGA-Net (Layer 3) & $62.92 \pm 0.11$ \\
MGA-Net (Layer 4) & $61.30 \pm 0.09$ \\
LLaVA -- Short Prompt & $34.96 \pm 0.03$ \\
LLaVA -- Long Prompt & $33.79 \pm 0.01$ \\
\bottomrule
\end{tabular}
\caption{
Additional Experiment Results - Comprehension Accuracies}
\label{table: Comp-acc-add}
\end{table}
For LLaVa \cite{llava}, we experimented with the following two prompts: 
\begin{itemize}
\item \textbf{Short prompt}: (USER: \textless image\textgreater\textbackslash n Give the bounding box for: "{Referring Expression}"\textbackslash nASSISTANT:)
\item \textbf{Long prompt}: (USER: \textless image\textgreater\textbackslash n Provide the bounding box coordinates for the object described by the referring expression: "{Referring Expression}"\textbackslash n ASSISTANT:)
\end{itemize}

Both prompts are similar in structure, but the latter prompt is more verbose.

For MGA-Net \cite{mganet}, we experiment with the Resnet-101 backbone by experimenting with Layers 3 and 4 for the visual features. 

The results of these experiments are in Table \ref{table: Comp-acc-add}. For LLaVA, the shorter prompt gives a slightly better result. For MGA-Net, the features captured by Layer 3 give a better result. Hence, we consider these two variants in all our experiments.

\noindent\paragraph{MGA-Net Hyperparameters.} Since MGA-Net is the only model we train in this paper, we provide the hyperparameters used. These hyperparameters are derived from \cite{mganet}'s work. The model is trained for 15 epochs, as validation performance begins to degrade beyond that point. Training uses the Adam optimizer with a learning rate of 1e-4, batch size of 30, and gradient clipping set to 0.3. The language encoder is a 2-layer Bi-LSTM with a hidden size of 512 and no dropout. Word embeddings are 512-dimensional. Visual features are extracted via ResNet101, and object features include both visual and normalized spatial information, processed through MLPs. The model employs Gated Graph Neural Networks (GGNNs) for multi-step relational reasoning, using 3 update steps.

\noindent\paragraph{Model parameters.}Among the models evaluated, \textbf{LLaVA} is a general-purpose vision-language model that integrates an open-set vision encoder from CLIP with a language decoder. The model is trained end-to-end through general visual instruction tuning to align visual and linguistic modalities. The commonly used LLaVA variant based on LLaMA-7B consists of approximately 7 billion parameters. \textbf{Grounding DINO} (base variant) has around 188 million parameters, optimized for phrase grounding and open-vocabulary detection. \textbf{DeepSeek-VL2 Tiny} is a compact model with about 600 million parameters, balancing speed and performance for multimodal tasks. \textbf{Qwen-VL 2.5} builds on the Qwen2.5 architecture and has 7 billion parameters, suitable for complex visual-language understanding. \textbf{OWL-ViT}, depending on the ViT backbone, ranges from \textbf{87M (ViT-B)} to \textbf{300M+ (ViT-L)} parameters, designed for open-vocabulary object detection. Finally, \textbf{MGA-Net}, tailored for referring expression grounding on datasets like CLEVR-Ref+, is lightweight with only 15--20 million parameters, yet delivers competitive task-specific performance.

\section{Experiments with other VLMs}\label{app: other models}
In our analysis, we also experimented with InstructBLIP (\citealp{instructblip}) and OpenFlamingo (\citealp{openflamingo}) models for the REC task. These models are general-purpose VLMs with InstructBLIP working in the zero-shot model and OpenFlamingo in the few-shot mode. In this section, we discuss the prompts that we used for these two models and the outputs obtained for the prompts:

\subsection{InstructBLIP}
For InstructBLIP, we designed three prompts for the REC task. They are as follows:
\begin{enumerate}
    \item Bounding Boxes: {bounding box list}; Referring Expression: {Refexp}; The index of the output bounding box is:
    \item Bounding Boxes: {bounding box list}; Referring Expression: {Refexp}; The coordinates of the output bounding box are:
    \item Provide the bounding box coordinates for: "{Refexp}"
\end{enumerate}
In the prompts, the ‘bounding box list’ placeholder takes the coordinates of the detected bounding boxes in the image being passed as the input, along with indices for each bounding box, starting from ‘1’. But for the third prompt, the model has no access to pre-detected candidate bounding boxes in the image. While the expected output for the first prompt is the index of the correct bounding box, for the other 2 prompts it is the bounding box coordinates as the output.

The bounding box format is [x1, y1, x2, y2], where (x1, y1) is the bottom left corner and (x2, y2) is the top right corner of the box. The coordinate values are a fraction of the total length/width of the image according to the position of the coordinate.

Unfortunately, none of the prompts gave consistently correct outputs. The outputs were as follows: \\
\textbf{Prompt 1}: The outputs were mostly incorrect. Sometimes, the model also gave ‘0’ as the output, even though it is not a valid index. \\
\textbf{Prompt 2}: The output did not return meaningful coordinates in most cases. But in the few instances that it did, they were mostly incorrect. Example outputs when the model could not return meaningful coordinates are: 
\begin{itemize}
  \item \texttt{\{1: [0.16, 0.55], 2: [0.32, 0.47], 3: [0.55, 0.6], 4: [0.21, 0.06]\}}
  \item \texttt{[0.9, 0.53, 0.93, 0.57, 0.0, 0.39]}
\end{itemize}

\textbf{Prompt 3}: The model could not understand the task, and it just paraphrased parts of the prompt instead of giving the coordinates as the output. Example prompts and outputs are:
\begin{itemize}
\item \textbf{Prompt:} Provide the bounding box coordinates for: "The large poster that is leaning against the wall" \\
\textbf{Output}: what is the bounding box coordinates for the large poster that is leaning against the wall
\item \textbf{Prompt:} Provide the bounding box coordinates for: "The young man that is leaning against the wall" \\
\textbf{Output}: is standing in an elevator. the young man that is leaning against the wall is standing in an elevator
\end{itemize}

\subsection{OpenFlamingo}
We tested all the prompts designed for OpenFlamingo in both 2 and 3-shot settings. \\\\
\textbf{Prompt 1}:
\begin{itemize}
    \item \textbf{Example output format:} \textless image\textgreater Bounding Boxes:{bounding box list}; Expression: {Refexp}; Correct Bounding Box:"ID"\textless\textbar endofchunk\textbar\textgreater
    \item \textbf{Query format:} \textless image\textgreater Bounding Boxes:{bounding box list}; Expression: {Refexp}; Correct Bounding Box:“
\end{itemize}
‘bounding box list’ placeholder takes the list of candidate bounding boxes in the image as input, in the same format as InstructBLIP (discussed in the previous section). The expected output is the index of the correct bounding box. However, we observed that irrespective of the query, the model gave the same output index for the same set of prompting examples. \\\\
\textbf{Prompt 2}: 
\begin{itemize}
    \item \textbf{Example output format:} \textless image\textgreater Expression: {Refexp}; Correct Bounding Box:[Bounding box coordinates]\textless\textbar endofchunk\textbar\textgreater
    \item \textbf{Query format:} \textless image\textgreater Expression: {Refexp}; Correct Bounding Box:[
\end{itemize}
‘bounding box list’ placeholder takes the same input as explained for Prompt 1. But instead of expecting the index, we expect the coordinates of the bounding box as the output. The format of the bounding box is the same as explained for InstructBLIP in the previous section. However, the model failed to give meaningful coordinates as output in most cases. When it did give meaningful coordinates, the outputs were mostly incorrect.

\end{document}